\documentclass[sigconf]{acmart}
\AtBeginDocument{%
  }
\renewcommand\footnotetextcopyrightpermission[1]{}
\usepackage{algorithm}
\usepackage{algorithmic}
\usepackage{enumitem}
\usepackage{booktabs}
\usepackage{graphicx}
\usepackage{subcaption}
\usepackage{array}
\usepackage{xcolor}
\usepackage{hyperref}
\usepackage{booktabs}
\usepackage{fontawesome5}

\begin{document}

\title{MegaTrain: Full Precision Training of 100B+ Parameter Large Language Models on a Single GPU\\ \vspace{5pt}{\large \faGithub\ \textbf{Code:} \url{https://github.com/DLYuanGod/MegaTrain}}}

\author{\vspace{-5pt}Zhengqing Yuan$^{\text{1}}$\enskip Hanchi Sun$^{\text{2}}$\enskip Lichao Sun$^{\text{2}}$\enskip Yanfang Ye$^{\text{1}}$\vspace{5pt}}
\affiliation{\vspace{1mm} $^{\text{1}}$University of Notre Dame \enskip $^{\text{2}}$Lehigh University \country{}}


\begin{abstract}
We present MegaTrain, a memory-centric system that efficiently trains 100B+ parameter large language models at full precision on a single GPU.
Unlike traditional GPU-centric systems, MegaTrain stores parameters and optimizer states in host memory (CPU memory) and treats GPUs as transient compute engines. 
For each layer, we stream parameters in and compute gradients out, minimizing persistent device state.
To battle the CPU-GPU bandwidth bottleneck, we adopt two key optimizations.
1) We introduce a pipelined double-buffered execution engine that overlaps parameter prefetching, computation, and gradient offloading across multiple CUDA streams, enabling continuous GPU execution.
2) We replace persistent autograd graphs with stateless layer templates, binding weights dynamically as they stream in, eliminating persistent graph metadata while providing flexibility in scheduling.
On a single H200 GPU with 1.5TB host memory, MegaTrain reliably trains models up to 120B parameters. It also achieves 1.84$\times$ the training throughput of DeepSpeed ZeRO-3 with CPU offloading when training 14B models. 
MegaTrain also enables 7B model training with 512k token context on a single GH200. 
\end{abstract}

\maketitle

\section{Introduction}

As Large Language Models (LLMs) scale to hundreds of billions of parameters~\citep{ye2025llms4all,team2025kimi}, the center of innovation is shifting from pretraining toward post-training regimes—instruction tuning, alignment, domain adaptation, and agent specialization~\citep{lai2025survey,tie2025survey}. Unlike trillion-parameter pretraining, these workloads are lightweight in computation and could, in principle, be performed on a single node~\citep{yuan2025efficientllm}. However, fine-tuning still requires loading full model parameters and optimizer states into memory, rendering hundred-billion-parameter models inaccessible on commodity hardware~\citep{yu2025differentially}. Meanwhile, GPU resources remain scarce: recent surveys indicate that \textbf{among 167 U.S. universities, only two achieve an average availability of more than one H100 GPU per student}~\citep{gpus_per_student_2025}. This scarcity creates a fundamental mismatch—while LLM development is transitioning toward memory-bound, node-scale post-training, most practitioners lack the GPU resources to participate.

One underutilized resource in current training systems is the memory hierarchy. Modern computer systems are equipped with multiple memory tiers—closer memory (e.g., HBM) being smaller, faster, and more expensive, while distant memory (e.g., DDR) is larger, slower, and cheaper—and data is placed across tiers according to access patterns to balance performance and cost. While offloading techniques such as ZeRO-Offload~\citep{rajbhandari2020zero} and ZeRO-Infinity~\citep{rajbhandari2021zero} have begun to extend GPU capacity by migrating model states to host memory (CPU memory) and NVMe storage, they do not fully exploit the memory hierarchy: the GPU remains the host of model parameters, while CPU and storage serve only as temporary spill buffers. In training, parameters, gradients, and optimizer states are accessed infrequently relative to activations, yet they remain persistently pinned in device memory (GPU memory). As model sizes grow, this underutilization of the memory hierarchy makes it increasingly difficult for end users, small companies, and research labs to train large models on their own hardware.

\begin{figure}[t]
    \centering
    \includegraphics[width=0.9\linewidth]{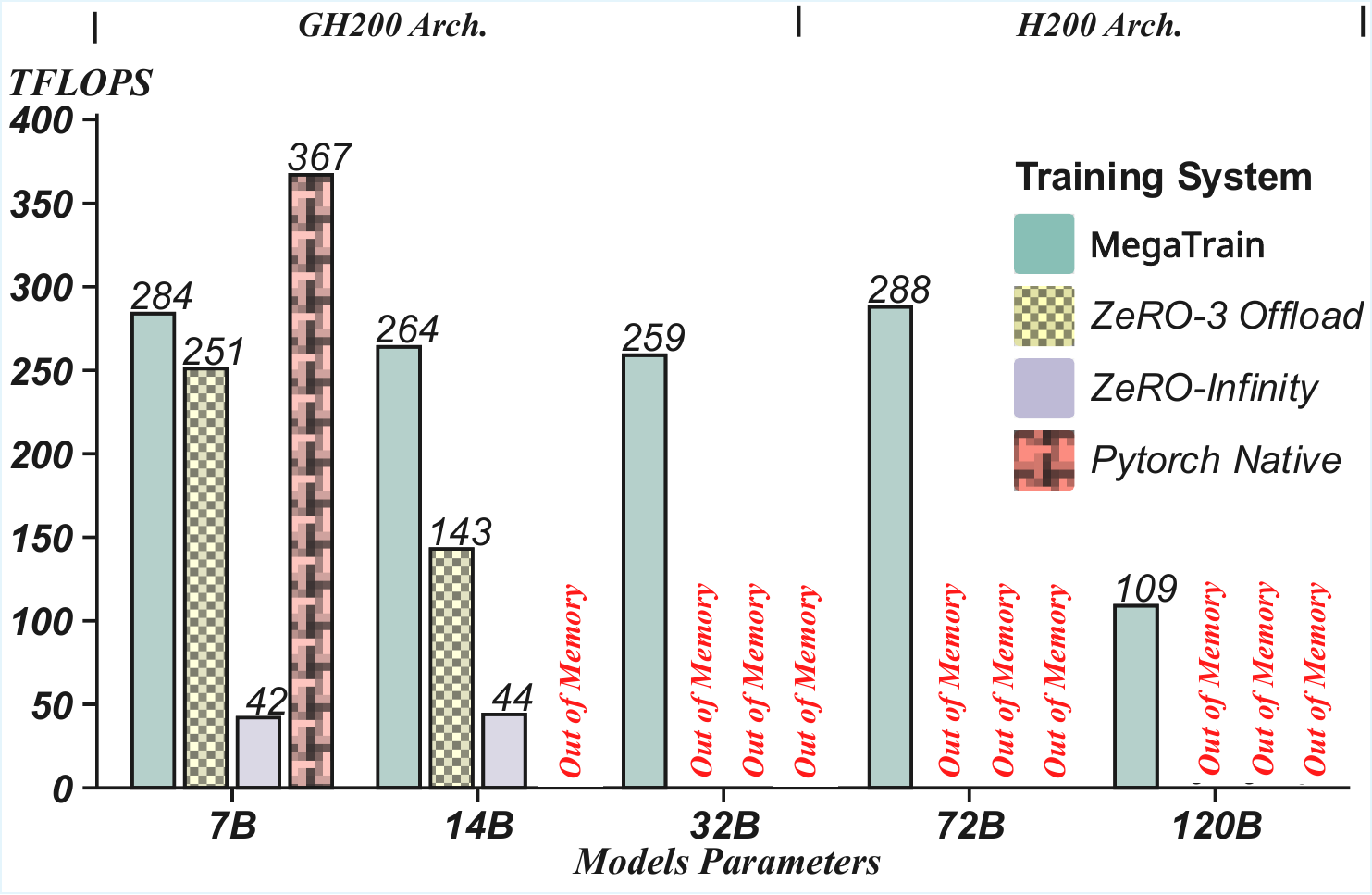}
    \caption{Sustained TFLOPS across model scales. MegaTrain remains efficient while offloading baselines become GPU memory bound.}
    \Description{Line plot of sustained training throughput across model scales on GH200 and H200. MegaTrain stays high and stable while the offloading baselines decline and eventually fail at larger model sizes.}
    \vspace{-12pt}
    \label{fig:floats}
\end{figure}

We present MegaTrain, a memory-centric training system that enables full precision training of 100B+ parameter large language models on a single GPU without sacrificing training speed. Instead of the traditional GPU-centric paradigm which mainly utilizes device memory, MegaTrain puts the parameters and optimizer states in host memory and uses GPUs only as transient compute engines equipped with a higher level cache (i.e. HBM/GDDR). During forward pass, parameters are streamed into GPU buffers on demand and immediately released after computation. During backward pass, the parameters are uploaded again while gradients are computed on the GPU and then streamed back to host memory. We keep the intermediate activations in GPU buffers and adopt a block-wise recomputation strategy to avoid the accumulation of activation memory. This design makes I/O independent of token count, allowing the system to use large batch size that amortizes the fixed I/O cost. All optimizer states remain in host memory and are updated using the CPU to avoid device memory pressure. Such design decouples the model scale from device memory size, making it possible to train large models on a single GPU with limited memory.

To battle the CPU-GPU bandwidth bottleneck, we adopt two key optimizations. 1) We introduce a pipelined double-buffered execution engine that overlaps parameter prefetching, computation, and gradient offloading across multiple CUDA streams, enabling continuous GPU execution while keeping memory footprints bounded. 2) Autograd builds a global computation graph that assumes all parameters and intermediate activations will stay parked in device memory until the entire backward pass is finished. If you are streaming weights in and evicting them layer-by-layer, that assumption breaks. By switching to a stateless template binding model, you decouple the math operations from persistent data. The GPU holds an empty "template" (for Transformers Layer) and dynamically binds it to whatever weights stream in. This completely eliminates the need to store the massive graph metadata and persistent intermediate tensors, guaranteeing that device memory usage never exceeds the footprint of a single layer. The decoupling also enables easier streaming of gradients back to host memory. With numerous other optimizations, we reach a higher training throughput and flops utilization than existing offloading-based systems.

In our evaluation, we demonstrate that MegaTrain can train models up to 120B parameter scale on a single H200 GPU equipped with 1.5TB host memory, a regime that existing offloading-based systems fail to reliably support.
On a single GH200, MegaTrain achieves 1.84$\times$ the training throughput of ZeRO-3 Offload at 14B scale and sustains over 250 TFLOPS at 32B, where existing offloading baselines encounter out-of-memory failures.
As a byproduct of its layer-wise memory design, MegaTrain also supports ultra-long context training up to 512K tokens on a single GH200.

\section{Preliminaries}

\subsection{Training Memory}
\label{sec:training-memory}

Memory usage when training large language models can be roughly categorized into three parts: 1) persistent state (parameters, gradients, optimizer states), 2) activations, 3) operator workspaces.
\textbf{Persistent state}: For mixed-precision training with Adam Optimizer~\citep{micikevicius2018mixed, kingma2015adam}, for each parameter, memory needs to store BF16 weights (2\,B), BF16 gradients (2\,B), and FP32 optimizer moments ($m$ and $v$, 8\,B combined), giving a minimum of $12P$ bytes for $P$ parameters.
For a 70B model, this requires at least 840\,GB of persistent state.
\textbf{Activations}: During backward pass, gradient calculation often requires intermediate activations, typically stored during forward pass. Yet, activation memory usage varies with the recomputation strategy and gradient calculation method. In MegaTrain, we adopt a block-wise recomputation strategy, storing only one checkpoint of activations every $K$ layers. This reduces the memory usage of activations to $O(N \cdot A_{\max} \cdot \frac{L}{K})$ bytes, where $N$ is the number of tokens in a batch, $A_{\max}$ is the maximum activation size of any single layer and $L$ is the model depth.
\textbf{Operator workspaces}: The workspace memory usage is hard to bound, but we assume it is bounded by a constant $W_{\max}$ bytes.

\subsection{Memory Hierarchy}
\label{sec:memory-hierarchy}

Modern GPU servers expose a four-tier memory hierarchy (Table~\ref{tab:memory-hierarchy}) that follows the classic fast-expensive-small to slow-cheap-large trade-off~\citep{hennessy2011computer}.
\textbf{On-chip SRAM} provides the highest bandwidth (${\sim}80$\,TB/s) but only tens of megabytes of capacity.
\textbf{Device memory} (GPU memory, e.g., HBM or GDDR) offers tens to hundreds of gigabytes at terabytes-per-second bandwidth, serving as the primary compute memory inside the GPU.
\textbf{Host memory} (CPU memory, e.g., DDR or LPDDR) supplies terabytes of capacity at hundreds of gigabytes per second, at roughly $10\times$ lower cost per byte than HBM.
\textbf{NVMe SSDs} add tens of terabytes of persistent storage at single-digit GB/s bandwidth.

This work focuses on the \emph{CPU--GPU boundary}, which differs significantly across server setups.
\begin{table}[t]
    \centering
    \caption{Memory hierarchy of H200 and GH200.}
    \small
    \setlength{\tabcolsep}{3pt}
    \resizebox{\columnwidth}{!}{
    \begin{tabular}{@{}lccc@{}}
    \toprule
    Tier & Capacity & Bandwidth & Unit Cost (\$/GB) \\
    \midrule
    SRAM (on-chip)            & $\sim$50--112\,MB & $\sim$80\,TB/s  & --- \\
    \addlinespace[3pt]
    \multicolumn{4}{@{}l}{\textit{H200 SXM}} \\
    \quad HBM3e (per GPU)     & 141\,GB      & 4.8\,TB/s       & $\sim$20 \\
    \quad Host DDR5           & 2--4\,TB     & $\sim$200\,GB/s  & $\sim$5--12 \\
    \quad PCIe Gen5 link & ---  & 128\,GB/s  & --- \\
    \addlinespace[3pt]
    \multicolumn{4}{@{}l}{\textit{GH200}} \\
    \quad HBM3 (per GPU)      & 96\,GB       & 4.0\,TB/s       & $\sim$20 \\
    \quad Host LPDDR5X        & 480\,GB      & 512\,GB/s  & $\sim$6--8 \\
    \quad NVLink-C2C link & ---  & 900\,GB/s  & --- \\
    \addlinespace[3pt]
    NVMe SSD                  & 10+\,TB      & 5--14\,GB/s     & $\sim$0.1 \\
    \bottomrule
\end{tabular}
    }
    
    \vspace{-12pt}
    \label{tab:memory-hierarchy}
\end{table}
For example, the H200 SXM uses host DDR5---a standard optimized for high per-pin frequency to match modern CPU architectures---connected via PCIe Gen5 at 128\,GB/s.
The GH200 GPU instead co-packages 480\,GB of LPDDR5X---a mobile-originated standard with a wider bus and lower power consumption, achieving 384--512\,GB/s aggregate bandwidth.
It accesses this memory via NVLink-C2C at 900\,GB/s, a ${\sim}7\times$ interconnect advantage over PCIe that fundamentally changes what offloading patterns are practical.

The principle governing memory hierarchies is to place data according to access patterns so that frequently or imminently needed items reside in fast tiers, while cold data migrates to slower, cheaper storage. When this placement aligns with workload locality, the system approaches the speed of the fastest tier at the cost of the slowest.
Most existing training systems treat device memory as the primary working memory.
While some frameworks offload optimizer states or parameters to host memory~\citep{rajbhandari2020zero,rajbhandari2021zero,zhao2023pytorch,fang_you_gemini_colossalai}, they regard it as a spill buffer rather than a first-class store.
MegaTrain inverts this relationship. Host memory becomes the primary store for all persistent training state, while device memory serves only as a transient compute cache. In practice, persistent states live in DDR or LPDDR, while active layer computation uses HBM or GDDR.


\subsection{Execution Challenges}
\label{sec:exec-scheduling}


Fitting the model into memory is necessary but not sufficient; the system must also execute efficiently despite continuous data movement.
Standard autograd in frameworks like PyTorch builds a global computation graph and retains activations until the backward pass completes.
This design assumes all parameters live on the GPU throughout training---an assumption that breaks when weights stream in layer by layer and are evicted immediately after use.
CUDA Graphs offer low-overhead kernel replay but require a \emph{static} execution pattern; streaming introduces dynamic address bindings (ping-pong buffers), shifting synchronization points, and interleaved recomputation that cannot be captured in a single graph.
MegaTrain therefore adopts an explicit scheduling model that coordinates prefetch, compute, and gradient offload across multiple CUDA streams.

\section{MegaTrain System}
\label{sec:system}

In this section, we present the execution workflow and key system mechanisms that enable efficient training at 100B+ scale on a single GPU. We first describe the end-to-end execution workflow, detailing how MegaTrain performs forward and backward passes and what resides in versus what is transferred between host memory and device memory. We then present key system designs that address the CPU-GPU bandwidth bottleneck, including the streaming protocol, memory management, and double-buffering that overlaps asynchronous DMA transfers with GPU computation. Lastly, we discuss the stateless execution model that avoids storing the autograd graph in device memory. Algorithm~\ref{alg:horizon-workflow} summarizes the full training-step workflow, while Appendix~\ref{app:implementation} covers lower-level implementation details.

\begin{figure}[t]
    \centering
    \includegraphics[width=0.9\linewidth]{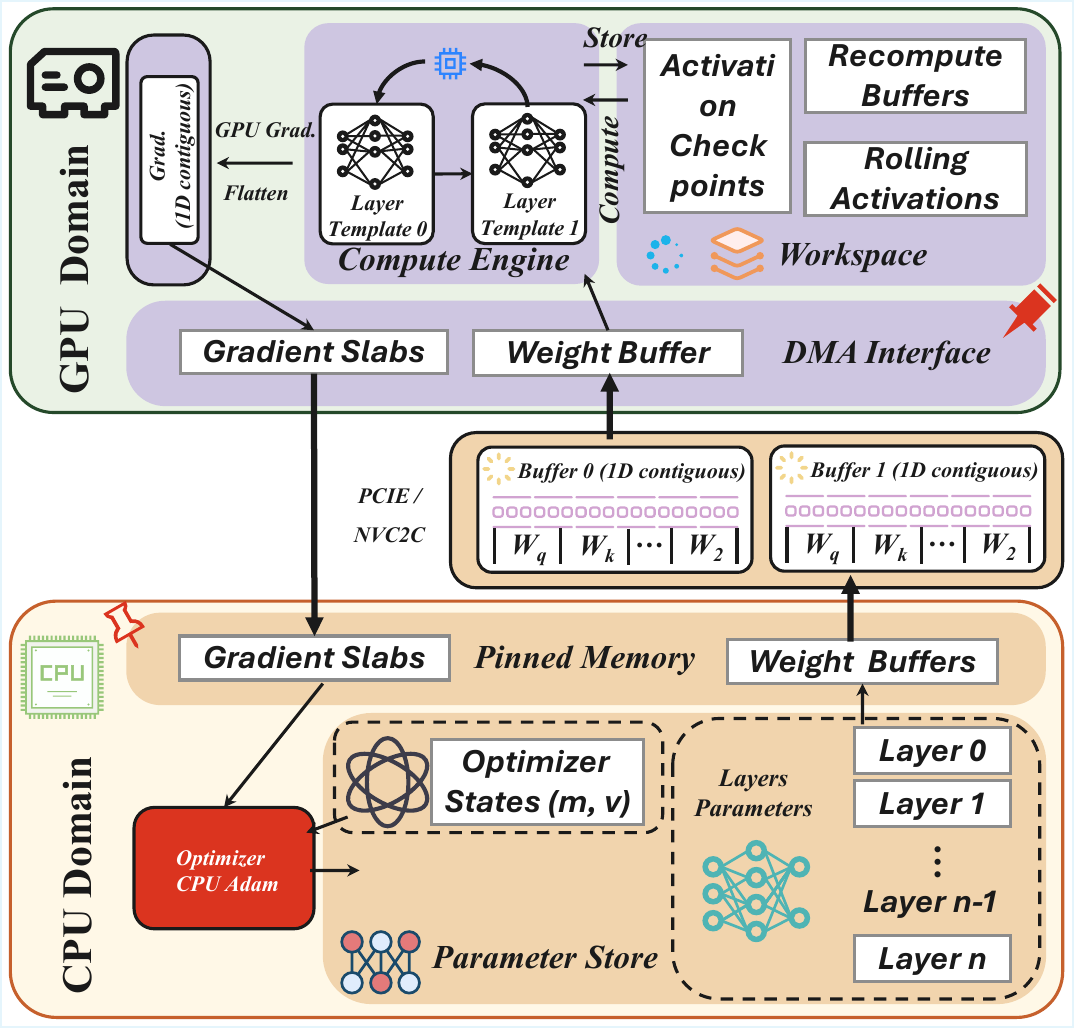}
    \caption{MegaTrain architecture: CPU serves as the parameter store while GPUs execute transient layer templates via asynchronous parameter streaming and gradient offloading.}
    \Description{System diagram showing Host (CPU) memory as the persistent store for parameters and optimizer states, and the GPU as a transient execution engine that streams weights in, computes each layer, and offloads gradients back to host memory.}
    \label{fig:overview}
    \vspace{-10pt}
\end{figure}

\subsection{Execution Workflow}
\label{sec:workflow}

As a memory-centric training system, MegaTrain possesses all persistent training states in host memory, including model parameters ($\theta$), optimizer states ($m, v$), and accumulated gradients. Device memory starts empty and only contains the layer template pool, which are lightweight, reusable compute kernels that dynamically bind to streamed parameters.


In the \emph{streaming forward} phase, we stream parameters from host memory into a weight buffer in device memory layer by layer. The buffer prefetches the next layer's weights so that the compute stream can immediately bind and execute the layer. Upon completion of each layer, the buffer is released and the next layer's weights are streamed in. During this phase, we also checkpoint the activation every $L/K$ layers and keep the checkpoint in device memory.

In the \emph{streaming backward} phase, we proceed in reverse block order. For each block, we start from the checkpoint (which resides in device memory) and stream parameters in forward order (e.g., $W_0, W_1, W_2$ of the block) to recompute activations. We then stream parameters in reverse order to compute backward passes, immediately offloading each layer's gradients (e.g., $G_3, G_2, G_1, G_0$) to host memory. This block-wise recomputation trades extra forward compute for bounded memory, as we only need to store activations for one block at a time, keeping device memory independent of total model depth.

Finally, the \emph{optimizer update} phase executes entirely on the CPU. Unlike forward and backward passes where GPU acceleration provides orders-of-magnitude speedup, optimizer updates (e.g., Adam) are compute-light yet I/O-intensive—each parameter update requires few arithmetic operations but must access the full parameter, gradient, and two momentum states. As observed by ZeRO-Offload~\citep{rajbhandari2020zero}, offloading optimizer computation to the CPU avoids the costly round-trip of streaming optimizer states to the GPU and back. Since PCIe bandwidth is the bottleneck rather than compute, executing Adam on the CPU with efficient vector instructions (e.g., AVX-512) matches or exceeds the throughput of GPU-based updates while eliminating four times the data movement.

\begin{algorithm}[t]
\caption{MegaTrain Training Step}
\label{alg:horizon-workflow}
\small
\begin{algorithmic}[1]
\REQUIRE Input batch $x$, parameters $\{\theta_i\}_{i=1}^L$ in host memory, checkpoint interval $K$
\ENSURE Updated parameters $\{\theta_i\}$
\STATE \textbf{// Phase 1: Streaming Forward}
\STATE $h_0 \leftarrow \mathrm{Embed}(x)$
\FOR{$i = 1$ to $L$}
    \STATE $\theta_i \leftarrow \mathrm{StreamIn}(i)$ \hfill \textit{// H2D transfer}
    \STATE $h_i \leftarrow f_i(h_{i-1}; \theta_i)$
    \IF{$i \bmod K = 0$}
        \STATE $\mathrm{Checkpoint}(h_i)$
    \ENDIF
    \STATE $\mathrm{Release}(\theta_i)$
\ENDFOR
\STATE \textbf{// Phase 2: Loss Anchoring}
\STATE $\ell \leftarrow \mathcal{L}(h_L)$; \quad $g_L \leftarrow \partial \ell / \partial h_L$
\STATE $\nabla \theta_{\mathrm{head}} \leftarrow \mathrm{BackwardHead}(\ell)$
\STATE $\mathrm{Offload}(\nabla \theta_{\mathrm{head}})$
\STATE \textbf{// Phase 3: Block-wise Backward}
\FOR{$b = \lfloor L/K \rfloor$ \textbf{downto} $0$}
    \STATE $h_{bK} \leftarrow \mathrm{LoadCheckpoint}(bK)$
    \STATE $\{h_j\}_{j=bK}^{(b+1)K} \leftarrow \mathrm{RecomputeBlock}(h_{bK})$
    \FOR{$i = (b+1)K$ \textbf{downto} $bK+1$}
        \STATE $\theta_i \leftarrow \mathrm{StreamIn}(i)$
        \STATE $(g_{i-1}, \nabla \theta_i) \leftarrow \mathrm{LocalBackward}(h_{i-1}, g_i; \theta_i)$
        \STATE $\mathrm{Offload}(\nabla \theta_i)$ \hfill \textit{// D2H transfer}
        \STATE $\mathrm{Release}(\theta_i)$
        \STATE $g_i \leftarrow g_{i-1}$
    \ENDFOR
\ENDFOR
\STATE \textbf{// CPU-side optimizer update (asynchronous)}
\STATE $\theta \leftarrow \mathrm{AdamUpdate}(\theta, \nabla\theta, m, v)$
\end{algorithmic}
\end{algorithm}





\subsection{Pipelined Execution Engine}

\begin{figure*}[t]
    \centering
    \includegraphics[width=0.9\linewidth]{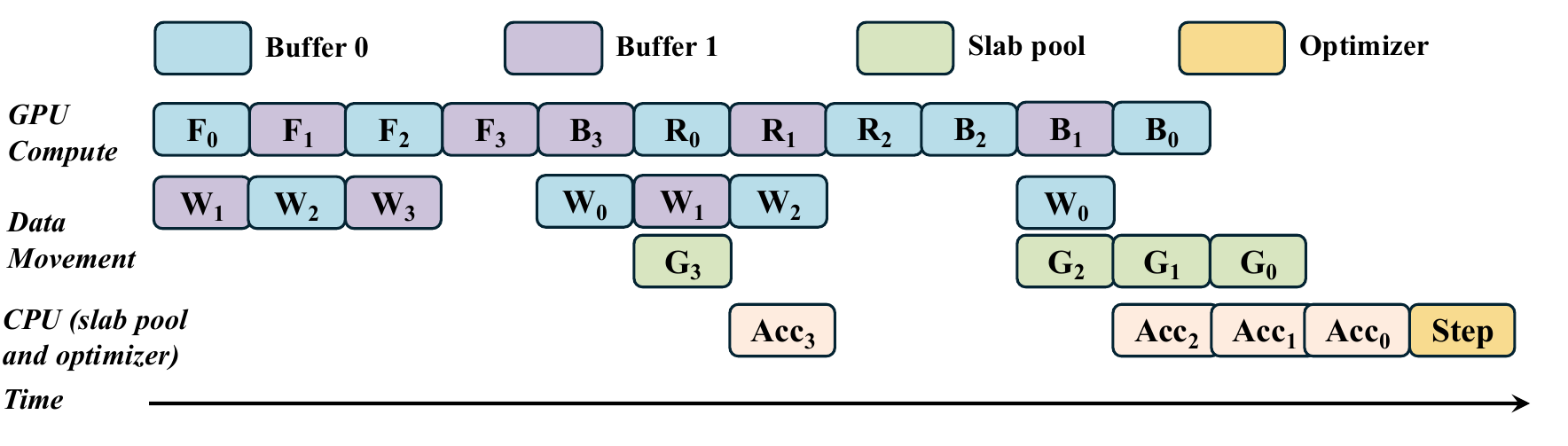}
    \vspace{-10pt}
    \caption{End-to-end pipelined execution. Weight prefetch ($W$), computation ($F$/$R$/$B$), and gradient offload ($G$) overlap across three CUDA streams.}
    \Description{Timeline diagram of three CUDA streams. Weight transfers, forward and recompute and backward compute, and gradient offloads overlap in a double buffered schedule so communication stays off the critical path.}
    \vspace{-10pt}
    \label{fig:pipeline}
\end{figure*}

The key to efficient streaming is hiding data movement latency behind computation. MegaTrain orchestrates three concurrent CUDA streams---compute ($S_{\mathrm{comp}}$), weight transfer ($S_{\mathrm{H2D}}$), and gradient transfer ($S_{\mathrm{D2H}}$)---with double-buffered staging to achieve continuous GPU execution (Figure~\ref{fig:pipeline}).

For streaming to be effective, the transfer time of each layer's parameters ($P_i / B_{\mathrm{pcie}}$) must be fully hidden under the preceding layer's computation; violating this overlap condition directly serializes execution. MegaTrain addresses this through \emph{double-buffered weight streaming}: it maintains two sets of staging buffers in both CPU and GPU domains, enabling a ping-pong prefetching strategy where the compute stream executes layer $F_i$ using Buffer~0 while the weight-transfer stream concurrently packs and streams layer $W_{i+1}$ into Buffer~1. This overlapping ensures that GPU compute units never stall waiting for parameters, converting sequential execution into a steady-state streaming pipeline. The same double-buffering applies during backward. While gradients from layer $i$ are being evacuated, parameters for layer $i-1$ are already streaming in.

Separating $S_{\mathrm{D2H}}$ from the compute stream is equally critical for asynchronous gradient evacuation. As shown in Figure~\ref{fig:pipeline}, gradient $G_3$ is evacuated while recomputation $R_0, R_1$ proceeds in parallel. By treating gradient offloading as a background task, MegaTrain prevents D2H latency from entering the critical path, ensuring GPU throughput is limited only by compute or H2D bandwidth.

Without a global training graph, the runtime cannot rely on implicit dataflow dependencies. Coordination across streams is governed by lightweight CUDA events: (1)~a \emph{Weights-Ready} event recorded by $S_{\mathrm{H2D}}$ after completing $W_i$, which $S_{\mathrm{comp}}$ waits on before binding layer~$i$; (2)~a \emph{Backward-Done} event recorded by $S_{\mathrm{comp}}$ after materializing $\nabla\theta_i$, which triggers $S_{\mathrm{D2H}}$ for evacuation; and (3)~a \emph{Buffer-Free} event recorded by $S_{\mathrm{D2H}}$ after offload completes, which $S_{\mathrm{H2D}}$ waits on before reusing the buffer (see Appendix~\ref{app:implementation}).

\subsection{Memory Management}

Conventional frameworks often manage tensors as fragmented objects scattered across the heap, forcing numerous small-granularity DMA requests that incur kernel launch overhead and PCIe tail-latency. MegaTrain instead employs layer-contiguous tiling: all states for each layer---BF16 weights, BF16 gradients, and FP32 Adam moments---are packed into a single contiguous block aligned to 4KB pages, enabling single-burst DMA transfers that saturate PCIe bandwidth (${\sim}26$\,GB/s on Gen4 x16).

Rather than pinning the entire model, which would exhaust host physical memory and page table resources, MegaTrain allocates a small pool of pinned staging buffers sized to $P_{\max}$. During \texttt{StreamIn}, a dedicated CPU worker performs JIT packing from the pageable layer store to a pinned slab, from which DMA transfers can proceed at full PCIe bandwidth. Double-buffering ensures this overhead overlaps with GPU execution, keeping the host-side pinning footprint constant regardless of model depth.

To avoid blocking compute on gradient offloading, MegaTrain maintains a pool of $K$ pinned host slabs for gradient evacuation. When a layer's backward completes, $S_{\mathrm{D2H}}$ immediately transfers gradients to an available slab. A background CPU thread monitors these slabs via \texttt{cudaEventSynchronize}, accumulating gradients into the master store and applying optimizer updates in parallel with GPU execution.


\subsection{Stateless Execution Model}

Standard autograd graphs assume parameters and activations persist on the GPU throughout backward propagation. Under layer-wise streaming, however, parameters are evicted after use and activations cannot be retained arbitrarily, rendering the global graph abstraction inapplicable. MegaTrain therefore adopts a \emph{stateless} execution model that decouples mathematical structure from physical data.

Rather than maintaining persistent model state on the GPU, MegaTrain employs a stateless template pool where each template (e.g., Template~A/B in Figure~\ref{fig:pipeline}) encapsulates the CUDA kernels for Attention and MLP blocks but holds no persistent weight pointers. Before execution, the \texttt{Bind} primitive dynamically maps views from the streaming buffer to the template's input slots. This ping-pong binding allows $F_1$ to execute on Template~A while $W_2$ is being bound to Template~B, eliminating weight preparation latency from the critical path.

MegaTrain also does not rely on CUDA graph capture. Because streamed weights, buffer ownership, and synchronization points change at layer granularity, the runtime preserves an explicit \texttt{StreamIn}-\texttt{Bind}-\texttt{Compute}-\texttt{Offload} dispatch path instead of forcing execution into a static captured graph. This explicit control enables dynamic buffer assignment and precise tensor lifecycle management, which is essential for the deterministic memory bound.

\section{Evaluation}
\subsection{Experimental Setup}
\textit{\textbf{GH200 System.}}
Experiments on the GH200 system are conducted on Grace-Hopper nodes. Each node contains four GH200 superchips, where each superchip integrates a 72-core Grace ARM CPU and one NVIDIA GH200 GPU with 96\,GB HBM3 memory, connected through NVLink-C2C with a peak bidirectional bandwidth of approximately 900\,GB/s. For evaluation, we use a single GH200 superchip with one GH200 GPU and 480\,GB of host memory from the local Grace CPU.

\textit{\textbf{H200 System.}}
We additionally evaluate MegaTrain on a single NVIDIA H200 SXM node equipped with one Intel Xeon Platinum 8558 CPU (96 cores total) and 1.5\,TB of host memory. The H200 GPU provides 141\,GB HBM3e memory and connects to the host via PCIe Gen4. 

\textit{\textbf{Dataset.}}
We evaluate model accuracy on the \textbf{\textit{MetaMathQA}} benchmark, a large-scale mathematical reasoning dataset comprising approximately 395,000 English math problem-answer pairs. MetaMathQA is constructed via data augmentation techniques over base reasoning benchmarks such as GSM8K and MATH, producing diverse multi-step math word problems with deterministic ground-truth answers. In our experiments, we randomly divide the dataset into \textbf{70\% training} (approximately 276,500 samples) and \textbf{30\% testing} (approximately 118,500 samples). We report \textit{exact-match accuracy}, defined as whether the model’s final predicted answer exactly matches the reference answer for each problem.

Table~\ref{tab:model_setup} summarizes the base models used throughout the evaluation. We report the total parameter count, transformer depth, hidden size, and FFN size to show the architectural range covered by our experiments. For GPT-OSS-120B, the hidden size and FFN size are written as per-expert width times the number of experts, which reflects its MoE design rather than a dense layout.

\begin{table}[t]
\centering
\footnotesize
\caption{Model configurations used in experiments.}
\label{tab:model_setup}
\resizebox{\columnwidth}{!}{
\begin{tabular}{@{}lcccc@{}}
\toprule
Model & Total Params & Layers & Hidden Size & FFN Size \\
\midrule
Qwen2.5-7B   & 7B   & 28 & 3584 & 18944 \\
Qwen2.5-14B  & 14B  & 48 & 5120 & 13824 \\
Qwen2.5-32B  & 32B  & 64 & 5120 & 27648 \\
Qwen2.5-72B  & 72B  & 80 & 8192 & 29568 \\
GPT-OSS-120B (MoE) & 120B & 36 & 2880$\times$128 & 2880$\times$128 \\
\bottomrule
\end{tabular}
}
\vspace{-10pt}
\end{table}


\subsection{Feasibility Boundary}
\begin{figure}[t]
    \centering
    \includegraphics[width=0.9\linewidth]{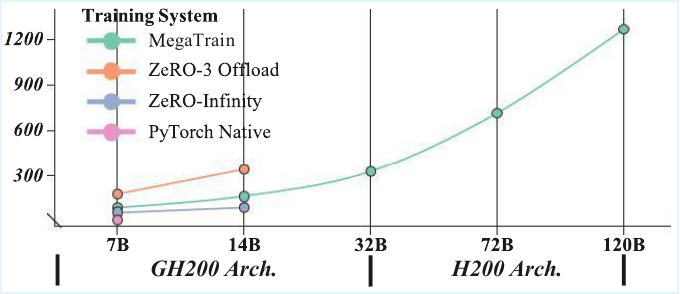}
    \vspace{-10pt}
    \caption{Host (CPU) memory footprint versus model scale across training systems.}
    \Description{Line plot of host memory usage versus model size for MegaTrain and baseline systems. MegaTrain grows close to linearly while the baselines rise faster and cross practical memory limits at smaller scales.}
    \label{fig:memory}
    \vspace{-12pt}
\end{figure}

All experiments in this subsection are conducted on two representative single-GPU platforms to illustrate how the feasibility boundary shifts with available host memory capacity. Models from 7B to 32B parameters are evaluated on a GH200 system, while larger models from 72B to 120B are evaluated on an H200 system equipped with 1.5\,TB host memory.

\textit{\textbf{Host Memory as the True Scaling Boundary.}}
The line plot in Figure~\ref{fig:memory} reports the host memory footprint required to train models of increasing scale under different training systems. A clear trend emerges. While ZeRO-3 Offload, ZeRO-Infinity, and PyTorch Native all exhibit rapidly increasing host memory consumption as model size grows, \textbf{MegaTrain} maintains a significantly flatter growth curve. From 7B to 120B parameters, competing systems show near-exponential growth in host memory demand due to redundant parameter staging, fragmented tensor storage, and optimizer state replication across offload buffers. In contrast, MegaTrain’s flat-tensor layout and authoritative CPU master storage ensure that memory growth is strictly proportional to the theoretical parameter footprint, without auxiliary duplication. This result highlights a critical feasibility boundary. For large models, \emph{host memory capacity}, rather than device memory, becomes the primary limiting factor for single-device training. Existing offloading systems cross this boundary rapidly beyond 30B parameters, while MegaTrain remains well within practical limits even at 120B scale.

\textit{\textbf{Compute Efficiency and Sustained TFLOPS.}}
Figure~\ref{fig:floats} reports sustained training throughput in TFLOPS across two architectures (GH200 and H200). At small scales (7B), PyTorch Native achieves high peak throughput due to full GPU residency, but this advantage collapses once models exceed device memory capacity. ZeRO-3 and ZeRO-Infinity suffer from substantial PCIe synchronization overhead and fragmented transfers, leading to severe degradation in sustained compute. MegaTrain, however, maintains consistently high TFLOPS across all scales. On GH200, MegaTrain sustains 284 TFLOPS at 7B, 264 TFLOPS at 14B ($1.84\times$ over ZeRO-3 Offload), and remains above 250 TFLOPS even at 32B. On H200, the system continues scaling to 72B and 120B while preserving high utilization. This stability arises from two design properties. First, large contiguous DMA transfers are enabled by pinned staging buffers. Second, compute and weight prefetch overlap is achieved through double buffering and stream execution.

\begin{table}[t]
\centering
\small
\caption{Final accuracy comparison across systems at 7B and 14B scales.}
\label{tab:acc_compare}
\begin{tabular}{lccccc}
\toprule
\textbf{Metric} & \textbf{Baseline} & \textbf{\shortstack{ZeRO-3\\Offload}} & \textbf{\shortstack{ZeRO\\Infinity}} & \textbf{\shortstack{PyTorch\\Native}} & \textbf{Ours} \\
\midrule
7B Acc. (\%)  & 33.47 & 88.93 & 88.97 & 88.91 & \textbf{88.99} \\
14B Acc. (\%) & 37.58 & 92.41 & 92.36 & - & \textbf{92.52} \\
\bottomrule
\end{tabular}
\end{table}

\textit{\textbf{Correctness Preservation at Scale.}}
Table~\ref{tab:acc_compare} shows that MegaTrain matches the numerical accuracy of standard full-GPU training and ZeRO-based baselines at both 7B and 14B scales. The negligible difference in accuracy confirms that MegaTrain’s explicit recompute and CPU-master design do not introduce numerical drift or optimization instability. This validates that the system’s memory and compute advantages do not trade off training correctness.

\subsection{Ablation Studies}
\begin{table}[t]
\centering
\small
\setlength{\tabcolsep}{4pt}
\caption{Ablation study of MegaTrain. Removing double buffering significantly degrades throughput, while other components have minor impact.}
\label{tab:ablation_megatrain}
\begin{tabular}{lccc}
\toprule
\textbf{Configuration} & \textbf{BS} & \textbf{TFLOPS} & \textbf{Device Mem (GB)} \\
\midrule
MegaTrain (full)         & 96 & 266.3  & 75.93 \\
w/o Double Buffering     & 96 & 182.91 & 74.11 \\
w/o Gradient Slab Pool   & 96 & 257.55 & 75.93 \\
w/ Checkpoint Interval=1 & 64 (96 OOM) & 240.45 & 81.34 \\
\bottomrule
\end{tabular}
\end{table}

Table~\ref{tab:ablation_megatrain} ablates MegaTrain's components to isolate their performance contributions.
Removing double buffering leads to a substantial drop in throughput, from 266.3 TFLOPS to 182.9 TFLOPS (a 31.3\% reduction), demonstrating that overlapping parameter prefetching, computation, and gradient offloading is critical for maintaining high GPU utilization. Without this mechanism, the GPU frequently stalls due to synchronization gaps between data transfer and compute.

In contrast, removing the gradient slab pool results in only a minor throughput decrease (266.3 $\rightarrow$ 257.6 TFLOPS), indicating that while memory pooling improves allocation efficiency and reduces fragmentation, it is not a primary driver of performance.

We evaluate the effect of recomputation granularity by setting the checkpoint interval to 1. This configuration significantly reduces the maximum feasible batch size (from 96 to 32) due to increased activation memory pressure, and lowers throughput to 184.2 TFLOPS. This highlights the importance of balancing recomputation frequency and memory usage to achieve optimal throughput.

\subsection{Depth Scalability Results}
\begin{figure*}[t]
    \centering
    \includegraphics[width=1\linewidth]{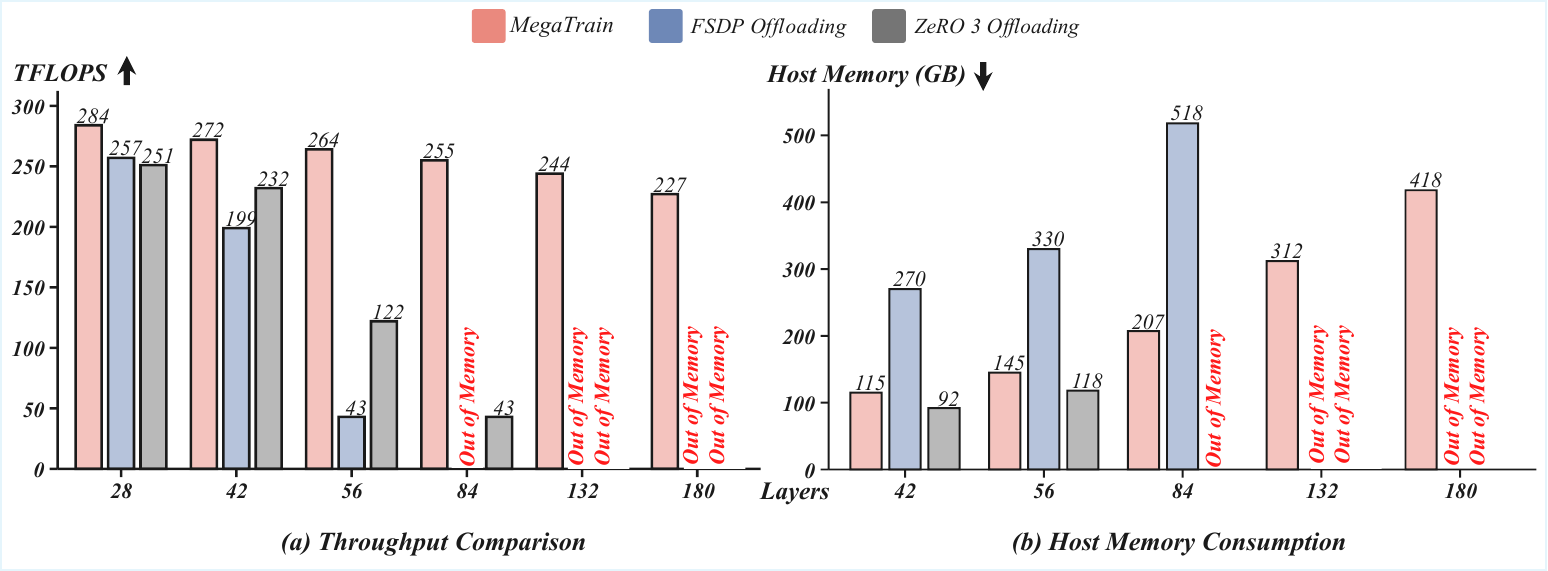}
    \vspace{-15pt}
    \caption{Depth scalability with fixed model width (hidden and FFN) size.}
    \Description{Two panel figure for depth scaling at fixed width. The left panel shows MegaTrain throughput staying comparatively stable as layer count increases while FSDP and ZeRO offloading degrade and fail. The right panel shows MegaTrain using less host memory than the baselines across the same models.}
    \label{fig:depth}
    \vspace{-10pt}
\end{figure*}

All experiments in this subsection are conducted on the GH200 system. Table~\ref{tab:depth_configs} and Figure~\ref{fig:depth}(a) and Figure~\ref{fig:depth}(b) evaluate how training systems behave when \emph{model depth} increases while \textbf{hidden size and device memory allocation remain strictly constant} (3.83\,GB). This setting isolates the system's capability to handle increasing parameter counts purely through depth scaling, without granting additional device memory. Such a setup directly stresses the memory orchestration, parameter movement, and recomputation efficiency of each system.

\begin{table}[t]
\centering
\footnotesize
\caption{Depth study model configurations.}
\label{tab:depth_configs}
\begin{tabular}{ccc}
\toprule
\textbf{Layers} & \textbf{Parameters (B)} & \textbf{Device Alloc (GB)} \\
\midrule
28  & 7.62  & 3.83 \\
42  & 10.88 & 3.83 \\
56  & 14.14 & 3.83 \\
84  & 20.67 & 3.83 \\
132 & 31.85 & 3.83 \\
180 & 43.04 & 3.83 \\
\bottomrule
\end{tabular}
\vspace{-10pt}
\end{table}

\textbf{\textit{Throughput under increasing depth.}}
As shown in Figure~\ref{fig:depth}(a), MegaTrain maintains remarkably stable throughput as depth increases from 28 to 180 layers. The throughput only decreases from 284\,TFLOPS to 227\,TFLOPS, a modest \textbf{20.1\%} drop despite the model growing from 10.9B to 43.0B parameters (a \textbf{3.95$\times$} increase in size). In contrast, both ZeRO-3 Offloading and FSDP Offloading exhibit severe throughput collapse as depth increases. At 42 layers, MegaTrain is already \textbf{1.37$\times$} faster than FSDP (272 vs.\ 199 TFLOPS) and \textbf{1.17$\times$} faster than ZeRO-3 (272 vs.\ 232 TFLOPS). At 56 layers, FSDP throughput drops catastrophically to 43 TFLOPS, making MegaTrain \textbf{6.14$\times$} faster (264 vs.\ 43). At 84 layers, ZeRO-3 degrades to 43 TFLOPS, where MegaTrain becomes \textbf{5.93$\times$} faster (255 vs.\ 43), while FSDP already runs out of memory. Beyond 84 layers, both baselines encounter OOM, whereas MegaTrain continues to scale to 132 and 180 layers with stable throughput. This demonstrates that existing offloading systems suffer from depth-induced communication and memory scheduling bottlenecks, where parameter movement and recomputation overhead grow superlinearly with depth.

\textbf{\textit{Host memory behavior.}}
Figure~\ref{fig:depth}(b) further reveals the host memory cost of enabling deeper models. At 42 layers, FSDP consumes 270\,GB host memory and ZeRO-3 uses 92\,GB, compared to only \textbf{115\,GB} for MegaTrain. At 56 layers, MegaTrain uses 145\,GB, while FSDP increases to 330\,GB (\textbf{2.28$\times$} higher). At 84 layers, FSDP reaches 518\,GB host memory before OOM, which is \textbf{2.50$\times$} higher than MegaTrain (207\,GB). At 132 and 180 layers, both baselines OOM due to host memory exhaustion, while MegaTrain continues operating at 312\,GB and 418\,GB respectively.

\begin{figure*}
    \centering
    \includegraphics[width=1\linewidth]{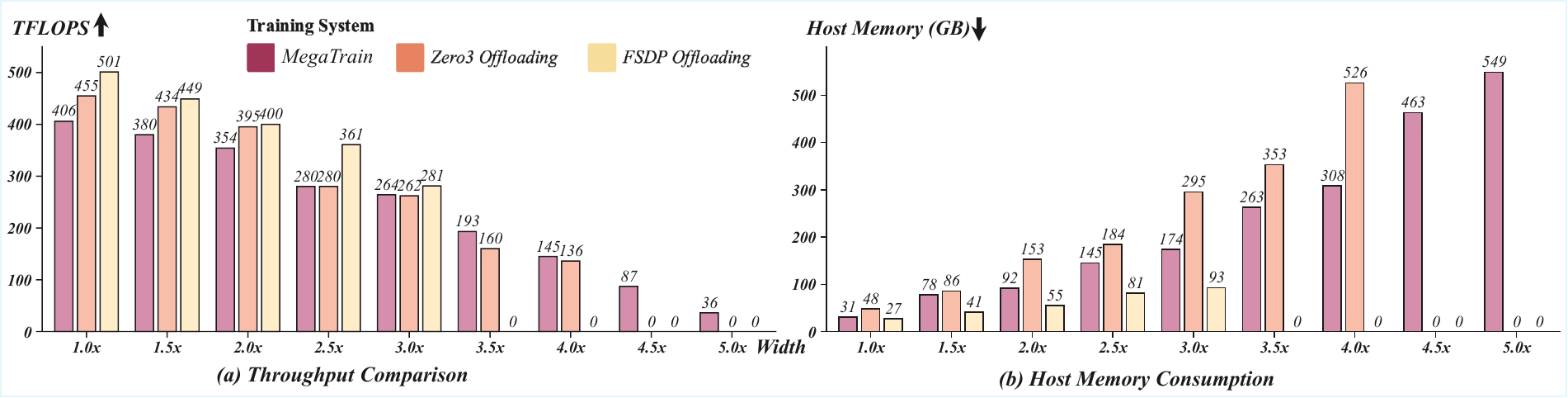}
    \caption{Width scalability with fixed model layers}
    \Description{Two panel figure for width scaling at fixed depth. Throughput decreases for all methods as width grows, but MegaTrain degrades more slowly and remains feasible at wider models. Host memory rises with width for all methods, with MegaTrain staying below the main offloading baselines.}
    \label{fig:width}
    \vspace{-5pt}
\end{figure*}

\subsection{Width Scalability Results}

Table~\ref{tab:width_configs} and Figure~\ref{fig:width} evaluate scalability when \emph{model width} (hidden and FFN dimensions) increases while the number of layers remains fixed. Unlike the depth experiment where GPU allocation remains constant, width scaling directly increases per-layer tensor sizes and therefore stresses device memory bandwidth, activation footprint, and parameter transfer volume. This experiment exposes a fundamentally different bottleneck compared to depth scaling.

\textbf{\textit{Throughput degradation with width.}}
All experiments in this subsection are conducted on the GH200 system.
As shown in Figure~\ref{fig:width}(a), all systems experience throughput reduction as width increases due to the quadratic growth of matrix multiplications. However, the \emph{rate of degradation} differs substantially. From 1.0$\times$ to 3.0$\times$ width, MegaTrain drops from 406 to 264 TFLOPS (\textbf{35.0\%} decrease). Over the same range, ZeRO-3 drops from 455 to 262 TFLOPS (\textbf{42.4\%} decrease). FSDP drops from 501 to 281 TFLOPS (\textbf{43.9\%} decrease).

\begin{table}[t]
    \centering
    \small
    \caption{Width study model configurations.}
    \label{tab:width_configs}
    \begin{tabular}{cccc}
    \toprule
    \textbf{Width Scale} & \textbf{Hidden Size} & \textbf{FFN Size} & \textbf{Device Alloc (GB)} \\
    \midrule
    1.0x & 3584  & 18944 & 3.83 \\
    1.5x & 5376  & 28416 & 7.01 \\
    2.0x & 7168  & 37888 & 11.07 \\
    2.5x & 8960  & 47360 & 15.99 \\
    3.0x & 10752 & 56832 & 21.78 \\
    3.5x & 12544 & 66304 & 28.44 \\
    4.0x & 14336 & 75776 & 35.97 \\
    4.5x & 16128 & 85248 & 44.36 \\
    5.0x & 17920 & 94720 & 53.62 \\
    \bottomrule
    \end{tabular}
    \vspace{-10pt}
\end{table}

Although MegaTrain starts slightly lower at small width (406 vs.\ 501 TFLOPS), its degradation curve is flatter. At 3.5$\times$ width, ZeRO-3 already falls to 160 TFLOPS while MegaTrain sustains 193 TFLOPS, making it \textbf{1.21$\times$} faster. At 4.0$\times$, ZeRO-3 further drops to 136 TFLOPS, where MegaTrain is \textbf{1.07$\times$} faster. Beyond 4.0$\times$, both ZeRO-3 and FSDP encounter OOM due to device and host memory pressure, while MegaTrain continues to operate up to 5.0$\times$ width.

\textbf{\textit{Host memory growth under width scaling.}}
Figure~\ref{fig:width}(b) shows that width scaling shifts pressure toward host memory due to increased size of parameter slabs and activation staging buffers. At 3.0$\times$ width, MegaTrain uses 174\,GB host memory, compared to 295\,GB for ZeRO-3 (\textbf{1.69$\times$} higher). At 3.5$\times$, ZeRO-3 reaches 353\,GB while MegaTrain uses 263\,GB (\textbf{1.34$\times$} higher). At 4.0$\times$, ZeRO-3 surges to 526\,GB and soon fails, while MegaTrain remains at 308\,GB. FSDP shows lower host memory at small widths but fails early (after 3.0$\times$) due to device memory fragmentation and activation pressure.

\subsection{Long Context Evaluation}

\begin{table}[t]
\centering
\small
\setlength{\tabcolsep}{4pt}
\caption{Long-context training performance on GH200. TFLOPS is computed using $6ND + 12LHT^2$. * denotes results with chunked MLP execution for ultra-long contexts.}
\label{tab:long_context}
\begin{tabular}{cccccc}
\toprule
\textbf{Context} & \textbf{BS} & \textbf{Tokens} & \textbf{Step (s)} & \textbf{TFLOPS} & \textbf{Mem} \\
\midrule
1K   & 158 & 162.7K & 27.05  & 284.7 & 74.2 GB \\
8K   & 25  & 204.8K & 32.36  & 294.5 & 86.5 GB \\
32K  & 6   & 196.6K & 32.18  & 316.7 & 84.0 GB \\
128K & 1   & 131.1K & 26.13  & 305.3 & 62.1 GB \\
256K & 1   & 262.1K & 236.1  & 401.2 & 88.2 GB \\
\midrule
512K$^*$ & 1 & 524.3K & 871.4 & 407.4 & 81.9 GB \\
\bottomrule
\end{tabular}
\vspace{-10pt}
\end{table}
Long-context training poses a distinct scaling challenge. Unlike model size scaling, which expands parameter storage, extending context length causes quadratic growth in attention computation and activation memory. Table~\ref{tab:long_context} reports MegaTrain's performance on a single GH200 as context length scales from 1K to 512K. TFLOPS improves consistently, rising from 264.8 at 1K to over 400 at 512K. Longer contexts increase arithmetic intensity from larger attention workloads, yielding better hardware utilization. Despite the 512$\times$ increase in sequence length, MegaTrain maintains stable memory usage and avoids activation explosion. Layer-wise execution limits activation residency to one layer at a time. For the extreme 512K setting, chunked MLP execution keeps memory within bounds without degrading throughput. These results show that MegaTrain supports ultra-long contexts while improving compute efficiency as context length scales.

\subsection{Verification on Different Devices}

\begin{table}[t]
\centering
\small
\caption{Nominal hardware characteristics of the three PCIe-based systems used in this subsection. Peak tensor throughput, PCIe bandwidth, and host memory bandwidth are nominal hardware specifications. PCIe bandwidth is the peak per-direction rate of an $\times$16 link.}
\label{tab:device_specs}
\resizebox{\columnwidth}{!}{
\begin{tabular}{@{}lccc@{}}
\toprule
 & \textbf{A100 PCIe} & \textbf{RTX A6000} & \textbf{RTX 3090} \\
\midrule
\textbf{Peak Tensor TFLOPS} & 312.0 & 154.8 & 142.0 \\
\textbf{Device Memory} & 80\,GB HBM2e & 48\,GB GDDR6 & 24\,GB GDDR6X \\
\textbf{Host Memory} & 600\,GB DDR4 & 251\,GB DDR4 & 251\,GB DDR4 \\
\textbf{PCIe BW} & 32\,GB/s & 32\,GB/s & 16\,GB/s \\
\textbf{Host BW} & 140.8\,GB/s & 204.8\,GB/s & 204.8\,GB/s \\
\bottomrule
\end{tabular}
}
\vspace{-10pt}
\end{table}

Table~\ref{tab:device_specs} summarizes the three PCIe-based systems used in this subsection. Together they span datacenter, workstation, and consumer GPUs, with device memory ranging from 24\,GB to 80\,GB and host-device links spanning PCIe Gen3 and Gen4.

\begin{figure}[t]
    \centering
    \includegraphics[width=0.9\linewidth]{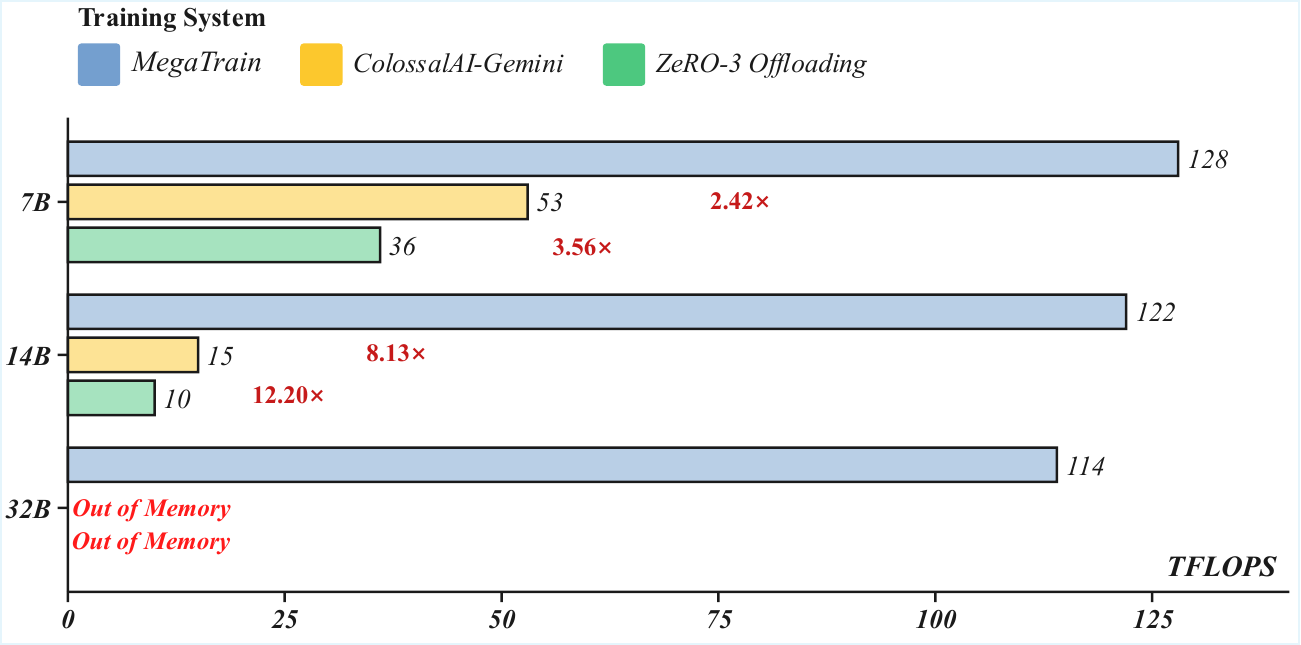}
    \caption{Performance comparison on a single A100 PCIe system.}
    \Description{Bar chart comparing achieved TFLOPS on a single A100 PCIe system for 7B, 14B, and 32B models. MegaTrain outperforms Gemini and ZeRO 3 across the supported scales and continues running when the baselines hit memory limits.}
    \label{fig:A100}
\end{figure}

\noindent\textbf{A100 System.}~~We first evaluate MegaTrain on a commodity PCIe-based server equipped with an Intel Xeon Platinum 8273CL processor, 600\,GB of host memory, and a single NVIDIA A100 GPU (80\,GB HBM2e) connected via PCIe Gen4. This setup represents a widely available datacenter configuration and shows that MegaTrain does not depend on NVLink-class interconnects. On this platform, we re-implement and tune two representative offloading baselines, ColossalAI-Gemini and ZeRO-3 CPU Offloading, following their official recommendations under the same software environment.

Figure~\ref{fig:A100} reports the achieved TFLOPS for 7B, 14B, and 32B models. Even on this different hardware stack, MegaTrain consistently outperforms both baselines. At 7B, MegaTrain reaches 128 TFLOPS, compared to 53 TFLOPS for Gemini and 36 TFLOPS for ZeRO-3, achieving \textbf{2.42$\times$} and \textbf{3.56$\times$} speedup respectively. At 14B, the gap widens. MegaTrain sustains 122 TFLOPS, while Gemini drops to 15 TFLOPS and ZeRO-3 to 10 TFLOPS, corresponding to \textbf{8.13$\times$} and \textbf{12.20$\times$} improvements. At 32B, both Gemini and ZeRO-3 encounter out-of-memory errors, while MegaTrain continues to operate at 114 TFLOPS.

\begin{table}[t]
\centering
\caption{Training performance comparison between MegaTrain and ZeRO-3 Offloading on A6000 (48GB) and RTX 3090 (24GB). 
ZeRO-3 results are reported with batch size (BS)=1; BS=2 leads to out-of-memory (OOM).}
\label{tab:training_perf}
\resizebox{\linewidth}{!}{
\begin{tabular}{lcccccc}
\toprule
\textbf{Model} & \textbf{Method} & \textbf{Max BS} & \textbf{TFLOPS} & \textbf{GPU Mem} & \textbf{CPU Mem} & \textbf{tok/s} \\
\midrule

\multicolumn{7}{c}{\textbf{A6000 (48GB VRAM)}} \\
\midrule
Qwen2.5-3B & MegaTrain & 15 & 49.70 & 46.74 GB & 38.3 GB & 2153 \\
           & ZeRO-3 Offload & 1 & 23.89 & 20.33 GB & -- & -- \\

Qwen2.5-7B & MegaTrain & 12 & 55.73 & 44.74 GB & 56.7 GB & 1219 \\
           & ZeRO-3 Offload & 1 & 27.55 & 20.83 GB & -- & -- \\

Qwen2.5-14B & MegaTrain & 9 & 56.82 & 44.64 GB & 104.1 GB & 641 \\
            & ZeRO-3 Offload & 1 & \textbf{OOM} & -- & -- & -- \\

\midrule
\multicolumn{7}{c}{\textbf{RTX 3090 (24GB VRAM)}} \\
\midrule
Qwen2.5-3B & MegaTrain & 7 & 33.18 & 22.83 GB & 25.0 GB & 1792 \\
           & ZeRO-3 Offload & 1 & 23.91 & 20.32 GB & -- & -- \\

Qwen2.5-7B & MegaTrain & 5 & 35.09 & 22.63 GB & 56.7 GB & 768 \\
           & ZeRO-3 Offload & 1 & 27.49 & 20.83 GB & -- & -- \\

Qwen2.5-14B & MegaTrain & 3 & 30.19 & 21.10 GB & 103.7 GB & 341 \\
            & ZeRO-3 Offload & 1 & \textbf{OOM} & -- & -- & -- \\

\bottomrule
\end{tabular}}
\end{table}

\noindent\textbf{Consumer-Grade Systems.}~~
To further evaluate MegaTrain on commodity hardware, we conduct experiments on a
single-node workstation equipped with an NVIDIA RTX A6000 (48\,GB, Ampere, PCIe Gen4)
and an NVIDIA GeForce RTX 3090 (24\,GB, Ampere, PCIe Gen3$\times$16).
Both GPUs share 251\,GB of host memory and are evaluated individually to
avoid PCIe bandwidth contention. All experiments use Qwen2.5 models with a fixed
sequence length of 8{,}192 and gradient checkpointing applied every 4 layers.

Table~\ref{tab:training_perf} summarizes the results. On the A6000, MegaTrain
scales up to 14B parameters, achieving 56.82 TFLOPS with batch sizes ranging from
9 to 15 depending on model size, while maintaining consistently high device memory
utilization (>93\%). On the RTX 3090, despite its limited 24\,GB device memory, MegaTrain
successfully trains 14B models at 30.19 TFLOPS, demonstrating strong scalability
on memory-constrained devices.

We further compare against ZeRO-3 with CPU offloading. While ZeRO-3 reduces device
memory usage, it is restricted to batch size 1 in our setting and fails to scale
to 14B models due to out-of-memory errors. Moreover, its throughput is
substantially lower (approximately 2$\times$ slower in TFLOPS) compared to
MegaTrain. These results highlight that MegaTrain not only enables larger batch
sizes and stable training at higher model scales, but also achieves significantly
higher hardware efficiency on consumer-grade GPUs.

\section{Conclusion}

We presented MegaTrain, a CPU-memory-centric training system that enables full-precision training of 100B+ parameter models on a single GPU. By treating host memory as the authoritative parameter store and GPUs as transient compute engines, MegaTrain decouples model scale from GPU memory capacity. The pipelined double-buffered execution engine overlaps parameter streaming with computation, while stateless layer templates eliminate the memory overhead of persistent autograd graphs. Together, these designs bound GPU memory to a per-layer footprint and allow host memory to scale linearly with model size.

More broadly, our work suggests that training large models is less about GPU capacity and more about memory and compute organization. When parameters stream through rather than persist, even commodity hardware can handle hundred-billion-parameter workloads. Extending MegaTrain to multiple GPUs with tensor or expert parallelism is a natural next step. Tiered storage that incorporates SSDs could push the boundary further, bringing trillion-parameter training within reach of everyday systems.


\bibliographystyle{ACM-Reference-Format}
\bibliography{sample-base}

\appendix
\section{Implementation Details}
\label{app:implementation}

MegaTrain is implemented as a high-performance training runtime leveraging PyTorch and CUDA. While the core logic resides in Python for flexibility, we offload critical path operations, such as batched parameter movement and SIMD-accelerated optimization, to C++ and CUDA extensions.

\subsection{Host (CPU) Memory Management}

MegaTrain treats host memory not merely as a secondary storage tier, but as the authoritative execution master. To support billion-parameter models on a single node, we design the CPU parameter store to maximize PCIe throughput and minimize host-side orchestration overhead.

\textit{\textbf{Layer-Contiguous Memory Tiling.}}
Conventional frameworks often manage tensors as fragmented objects scattered across the heap. This fragmentation forces the system to issue numerous small-granularity DMA requests, each incurring kernel launch overhead and PCIe transaction tail-latency.
To address this, MegaTrain implements \emph{Layer-Contiguous Tiling}. For each Transformer layer $i$, all associated states, including BF16 weights ($\theta_i$), BF16 gradients ($\nabla \theta_i$), and FP32 Adam moments ($m_i, v_i$), are packed into a single, monolithic memory block.
As shown in Algorithm~\ref{alg:horizon-workflow}, this layout ensures that the \texttt{StreamIn} primitive can satisfy a layer's residency transition with a single, large-burst DMA transfer, saturating the PCIe bus bandwidth. Furthermore, these tiles are aligned to 4KB page boundaries to facilitate zero-copy pinned staging.

\textit{\textbf{Pinned Slab Recycling.}}
A significant challenge in node-scale training is that pinning the entire model's parameters would exhaust host physical memory and OS page table resources.
MegaTrain employs a fixed-capacity \emph{Pinned Slab Pool} that acts as a staging area for the streaming engine. Instead of pinning the total model $L$, we only pin a small number of "active" slabs.
During the \texttt{StreamIn} phase, a dedicated CPU worker thread performs a JIT (Just-In-Time) copy from the pageable layer-contiguous store to a pinned slab. By double-buffering these slabs, we ensure that while the GPU executes layer $i$, the CPU is already packing and pinning layer $i+1$. This approach keeps the host-side pinning overhead constant ($O(P_{\max})$) regardless of model depth.

\textit{\textbf{Flat-Tensor Layout.}}
To eliminate the overhead of thousands of small PCIe transfers, MegaTrain enforces a \textbf{Flat-Tensor Layout}. During initialization, we extract the metadata (shape, numel) of all transformer layers and allocate two types of host-side memory:
\begin{itemize}[left=0pt]
    \item \textbf{Master Store:} Model parameters and FP32 Adam moments are stored in non-pinned host memory to maximize capacity.
    \item \textbf{Pinned Staging Buffers:} We allocate two fixed-size page-locked (pinned) buffers, each exactly matching the size of the largest transformer layer ($P_{\max}$). These buffers act as the H2D/D2H gateway, ensuring that all DMA transfers achieve near-peak PCIe bandwidth (e.g., $\sim$26GB/s on PCIe Gen4 x16).
\end{itemize}

\textit{\textbf{Structural Aliasing for Tied Weights.}}
To support models with tied embeddings, MegaTrain maintains a \emph{Virtual-to-Physical} mapping. Both the embedding and LM head point to the same physical memory tile in the CPU store. The execution controller tracks the "readiness" of this shared tile; once the head gradients are processed and the optimizer updates the shared parameters, the embedding is automatically marked as ready for the subsequent iteration's \texttt{StreamIn}, ensuring numerical consistency without redundant storage or synchronization barriers.

\textit{\textbf{Weight Tying.}} For models with tied Embedding and LM-Head, we implement \textit{aliased synchronization}. If the LM-Head and Embedding weights are tied, the system records the underlying \texttt{\detokenize{data_ptr}}. During the H2D sync phase, only one transfer is issued, and the pointers on the GPU are re-mapped to the same device memory address to prevent divergence.

\subsection{Multi-Stream Pipeline and Scheduling}
\label{app:scheduling}

The heart of MegaTrain is an event-driven scheduler that manages three concurrent CUDA streams: \texttt{ComputeStream}, \texttt{WeightStream}, and \texttt{GradStream}.

To saturate the PCIe bandwidth and maximize GPU utilization, MegaTrain implements a multi-stream pipeline that aggressively overlaps data movement with computation. As visualized in Figure~\ref{fig:pipeline}, the system orchestrates three concurrent hardware streams mediated by a hierarchy of CUDA events.

\textit{\textbf{Weight Double-Buffering.}}
For streaming to be effective, the transfer time of each layer's parameters ($P_i / B_{\mathrm{pcie}}$) must be fully hidden under the preceding layer's computation; violating this local overlap condition directly serializes execution regardless of available compute.
To eliminate the latency of parameter ingestion ($W_i$), MegaTrain maintains two sets of staging buffers in both the CPU and GPU domains. This enables a "ping-pong" prefetching strategy: while the \emph{compute stream} executes layer $F_i$ using \texttt{Buffer 0}, the \emph{weight-transfer stream} concurrently packs and streams layer $W_{i+1}$ into \texttt{Buffer 1}. As shown in Figure~\ref{fig:pipeline}, this overlapping ensures that the GPU compute units never stall for parameters, effectively converting a sequential execution into a steady-state streaming pipeline.

\textit{\textbf{Multi-Stream Orchestration.}}
MegaTrain separates execution into three dedicated CUDA streams to avoid false dependencies and global device synchronizations:
\begin{itemize}[left=0pt]
    \item \textbf{Compute Stream ($S_{\mathrm{comp}}$):} Responsible for executing the \texttt{Compute}, \texttt{RecomputeBlock}, and \texttt{LocalBackward} primitives.
    \item \textbf{Weight-Transfer Stream ($S_{\mathrm{H2D}}$):} Orchestrates asynchronous H2D copies of parameters $\theta_i$ (the $W_i$ blocks in Figure~\ref{fig:pipeline}).
    \item \textbf{Gradient-Transfer Stream ($S_{\mathrm{D2H}}$):} Manages the immediate evacuation of gradients $\nabla \theta_i$ to host-side gradient slabs (the $G_i$ blocks in Figure~\ref{fig:pipeline}).
\end{itemize}

\textit{\textbf{Event-Driven Synchronization.}}
Without a global training graph, the runtime cannot rely on implicit dataflow dependencies; it must explicitly track and enforce ordering across prefetch, computation, and offload operations.
The coordination across these streams is therefore governed by a lightweight event-driven protocol rather than heavy-weight host-side barriers:
\begin{enumerate}[left=0pt]
    \item \textbf{Weights-Ready Event:} Recorded by $S_{\mathrm{H2D}}$ after $W_i$ completes; $S_{\mathrm{comp}}$ waits on this event before invoking the \texttt{Bind} primitive for layer $i$.
    \item \textbf{Backward-Done Event:} Recorded by $S_{\mathrm{comp}}$ after the local gradient $\nabla \theta_i$ is materialized; this triggers $S_{\mathrm{D2H}}$ to initiate the evacuation $G_i$.
    \item \textbf{Buffer-Free Event:} Recorded by $S_{\mathrm{D2H}}$ after the gradient offload is finished. The $S_{\mathrm{H2D}}$ stream must wait for this event before reusing the corresponding buffer for the next iteration's weight prefetch.
\end{enumerate}

\textit{\textbf{Asynchronous Gradient Evacuation.}}
The separation of $S_{\mathrm{D2H}}$ from the compute stream is critical for maintaining throughput during the backward pass. As shown in Figure~\ref{fig:pipeline}, the evacuation of $G_3$ occurs in parallel with the recomputation $R_0$ and $R_1$. By treating the gradient return as a background task, MegaTrain prevents the PCIe D2H latency from leaking into the critical path of the backward recomputation, ensuring that the GPU's floating-point throughput is limited only by the slower of the compute kernel or the H2D parameter bandwidth.

\subsection{GPU Buffer Management}

The GPU domain functions as a transient execution cache, engineered to maximize throughput via a "just-in-time" parameter supply chain. As depicted in the Data Movement lane of Figure~\ref{fig:pipeline}, the engine provides high-bandwidth ingestion and evacuation while maintaining a stateless device profile.

\textit{\textbf{Flat-Buffer Streaming and Zero-Copy View.}}
To minimize the overhead of hundreds of individual CUDA API calls, MegaTrain employs a \emph{flat-buffer ingestion} strategy. For each layer $i$, the CPU packs all constituent tensors into a single contiguous pinned buffer (the $W_i$ blocks in Figure~\ref{fig:pipeline}). The \texttt{StreamIn} primitive issues a single asynchronous H2D copy. Upon arrival in the GPU's \texttt{Buffer 0/1}, the engine performs a \emph{zero-copy unflattening}: it creates tensor "views" that point directly into the flat buffer's offsets. This avoids repeated GPU-side memory allocations and ensures that parameter materialization occurs at near-line-rate PCIe speeds.

\textit{\textbf{Batched Parameter Binding.}}
To further reduce Python's dispatch overhead (which can exceed 10\% of step time), we developed a C++ extension for \textbf{Batched Parameter Binding}. This extension uses a single CUDA kernel to map the flattened GPU staging buffer back to the layer's named parameters (e.g., \texttt{\detokenize{q_proj}}, \texttt{\detokenize{k_proj}}) via pointer manipulation, replacing hundreds of individual \texttt{\detokenize{copy_}} calls with a single metadata update.

\textit{\textbf{K-Slab Gradient Offloading.}}
A naive gradient return would block the compute stream. MegaTrain introduces a \textbf{Categorized Gradient Slab Pool} consisting of $K$ pinned host memory "slabs" (default $K=12$).
When a layer's local backward pass completes, \texttt{GradStream} immediately issues a D2H transfer to an available slab. This decouples GPU memory release from CPU optimization. A dedicated background CPU thread monitors these slabs using \texttt{Event.synchronize()}, unflattening and accumulating them into the master store using OpenMP-parallelized kernels.

\textit{\textbf{Memory-Mapped Workspace Management.}}
Because the streaming regime requires precise control over what state is materialized and when it is released, MegaTrain replaces implicit graph-managed memory with explicit lifecycle control.
To further reduce runtime jitter, all transient workspaces for recomputation ($R_i$ in Figure~\ref{fig:pipeline}) and local activations are pre-allocated and memory-mapped at initialization. The engine manages these as a stack-like structure: the \texttt{RecomputeBlock} primitive pushes recomputed states onto this workspace, which are then popped and released by the \texttt{LocalBackward} primitive. This explicit lifecycle management guarantees that GPU memory fragmentation is zero, providing the deterministic $M_{\mathrm{GPU}}$ bound established in Section~\ref{sec:system}.

\textit{\textbf{Static Host-Side Staging.}}
To sustain the $S_{\mathrm{H2D}}$ and $S_{\mathrm{D2H}}$ streams without exhaustive pinning, MegaTrain partitions host memory into fixed-size regions.
First, two \emph{pinned staging buffers} (Buffer 0/1 in Figure~\ref{fig:pipeline}) are allocated to facilitate weight prefetching. This ensures the host-side pinning footprint remains invariant to model depth $L$.
Second, a \emph{slab pool} (the green blocks in Figure~\ref{fig:pipeline}) manages gradient returns. Slabs are recycled only after the CPU-side accumulation (\texttt{Acc}) completes, providing a back-pressure mechanism that prevents gradient offloading from overrunning host memory.

\textit{\textbf{Deterministic GPU Execution Cache.}}
The GPU domain is partitioned into a set of functional workspaces with strictly controlled lifetimes:
\begin{itemize}[left=0pt]
    \item \textbf{Streaming Buffers:} Dedicated buffers for the \texttt{StreamIn} primitive, sized to the maximum layer parameter volume $P_{\max}$.
    \item \textbf{Activation Stack:} A pre-allocated workspace for rolling activations and recomputation blocks. By managing this as a stack rather than a heap, MegaTrain avoids the fragmentation common in long-running training sessions.
    \item \textbf{Checkpoint Anchors:} A dedicated region for every $K$-th activation $h_{bK}$, which remains resident only until the corresponding block-wise backward pass is completed.
\end{itemize}

\textit{\textbf{Eliminating Runtime Jitter.}}
Beyond reducing allocator latency, this pooling strategy is critical for the robustness of the pipelined schedule shown in Figure~\ref{fig:pipeline}. By using pre-allocated, reusable buffers, MegaTrain eliminates "bubbles" in the pipeline caused by dynamic memory allocation or garbage collection. This architectural choice ensures that the system maintains a constant, high-throughput steady state, even when training hundred-billion-parameter models at the limit of the device's capacity.

\textit{\textbf{Fragmentation Control.}} We use the \texttt{\detokenize{expandable_segments}} flag in the PyTorch allocator to prevent virtual memory fragmentation during recomputation. By explicitly calling \texttt{\detokenize{record_stream}} on all transient buffers, we ensure that the allocator does not reclaim memory still in flight within the \texttt{GradStream}, avoiding silent data corruption.

\subsection{GPU Compute Dispatch}

\textit{\textbf{Stateless Template Binding.}}
MegaTrain decouples the layer's mathematical structure from its physical data through a \emph{stateless template pool}. Each template (e.g., Template~A/B) encapsulates the CUDA kernels for Attention and MLP blocks but possesses no persistent weight pointers. Before execution, the \texttt{Bind} primitive dynamically maps the views from the streaming buffer to the template's input slots. As visualized in the alternating colors of Figure~\ref{fig:pipeline}, this "ping-pong" binding allows $F_1$ to execute on Template~A while $W_2$ is being bound to Template~B, eliminating the latency of weight preparation from the critical path.

\textit{\textbf{Graph-less Dispatch.}}
Standard autograd graphs assume that parameters and activations persist on the GPU throughout backward propagation. Under layer-wise streaming, however, parameters are evicted after use and activations cannot be retained arbitrarily, rendering the global graph abstraction inapplicable.
MegaTrain therefore does not rely on CUDA graph capture. Because streamed weights, buffer ownership, and event dependencies change at layer granularity, the runtime preserves the explicit \texttt{StreamIn}-\texttt{Bind}-\texttt{Compute}-\texttt{Offload} dispatch path instead of forcing execution into a static captured graph.

\section{Ratel Reproduction on GH200}

We also tried to reproduce the Ratel~\citep{lohan} experiment on GH200 using the official codebase. The measured throughput is consistently low across all tested model sizes.

\begin{table}[t]
\centering
\footnotesize
\caption{Ratel reproduction results on GH200 using the official codebase.}
\label{tab:ratel_reproduction}
\begin{tabular}{lc}
\toprule
Model & TFLOPS \\
\midrule
7B  & 2.03 \\
14B & 10.90 \\
32B & 10.91 \\
\bottomrule
\end{tabular}
\vspace{-10pt}
\end{table}

We suspect that these low numbers are mainly caused by SSD bottlenecks in the original implementation.

\end{document}